\documentclass[10pt,twocolumn,letterpaper]{article}

\usepackage{cvpr}              

\usepackage{graphicx}
\usepackage{amsmath}
\usepackage{amssymb}
\usepackage{booktabs}
\usepackage{comment}
\usepackage[accsupp]{axessibility}
\usepackage{subcaption}
\usepackage{float}
\usepackage{caption,cuted}
\usepackage{appendix}
\newcommand*\samethanks[1][\value{footnote}]{\footnotemark[#1]}
\newcommand{\extrema}{bimodal\@\xspace}

\usepackage{multibib}
\newcites{append}{Appendix References}

\usepackage[pagebackref,breaklinks,colorlinks]{hyperref}

\usepackage[capitalize]{cleveref}
\crefname{section}{Sec.}{Secs.}
\Crefname{section}{Section}{Sections}
\Crefname{table}{Table}{Tables}
\crefname{table}{Tab.}{Tabs.}

\def\cvprPaperID{517} 
\def\confName{CVPR}
\def\confYear{2022}

\begin{document}

\title{Interactiveness Field in Human-Object Interactions\thanks{The research is supported in part by the Hong Kong Research Grant Council under grant number 16201420.}}

\author{Xinpeng Liu$^{1}$\thanks{The first two authors contribute equally.}~~~~~Yong-Lu Li$^{2}$\samethanks~~~~~Xiaoqian Wu$^{1}$~~~~~Yu-Wing Tai$^{3}$~~~~~Cewu Lu$^{1}$\thanks{Corresponding author.}~~~~~Chi-Keung Tang$^{2}$\\
$^{1}$Shanghai Jiao Tong University ~~~~~~~~~$^{2}$HKUST
~~~~~~~~~$^{3}$Kuaishou Technology\\
{\tt\footnotesize \{xinpengliu0907, yuwing\}@gmail.com, \{yonglu\_li, enlighten, lucewu\}@sjtu.edu.cn},
{\tt\footnotesize cktang@cs.ust.hk}
}

\maketitle

\begin{abstract}
   Human-Object Interaction (HOI) detection plays a core role in activity understanding. 
   Though recent two/one-stage methods have achieved impressive results, as an essential step, discovering interactive human-object pairs remains challenging. Both one/two-stage methods fail to effectively extract interactive pairs instead of generating redundant negative pairs.
   In this work, we introduce a previously overlooked \textbf{interactiveness \extrema prior}: given an object in an image, after pairing it with the humans,
   the generated pairs are either mostly non-interactive, or mostly interactive, with the former more frequent than the latter. 
   Based on this interactiveness \extrema prior we propose the \textbf{``interactiveness field''}. 
   To make the learned field compatible with real HOI image considerations, we propose new energy constraints based on the cardinality and difference in the inherent ``interactiveness field'' underlying interactive versus non-interactive pairs.
   Consequently, our method can detect more precise pairs and thus significantly boost HOI detection performance, which is validated on widely-used benchmarks where we achieve decent improvements over state-of-the-arts.
   Our code is available at  https://github.com/Foruck/Interactiveness-Field.
\end{abstract}
\vspace{-15px}
\section{Introduction}
\label{sec:intro}
Human-Object Interaction (HOI) detection consists of distinguishing human-object (H-O) pairs that have interactions from still images and classifying the interactions into various verbs. In practice, an HOI instance is represented as a triplet: $\langle \mathit{human}, \mathit{verb}, \mathit{object} \rangle$.
Considering its important role in recent advances in robot manipulation~\cite{hayes2017interpretable}, surveillance event detection~\cite{abnormal,unusualeventdetection}, and so on, HOI detection has been attracting continuous attention in computer vision. 

\begin{figure}[!t]
	\centering
	\includegraphics[width=0.5\textwidth]{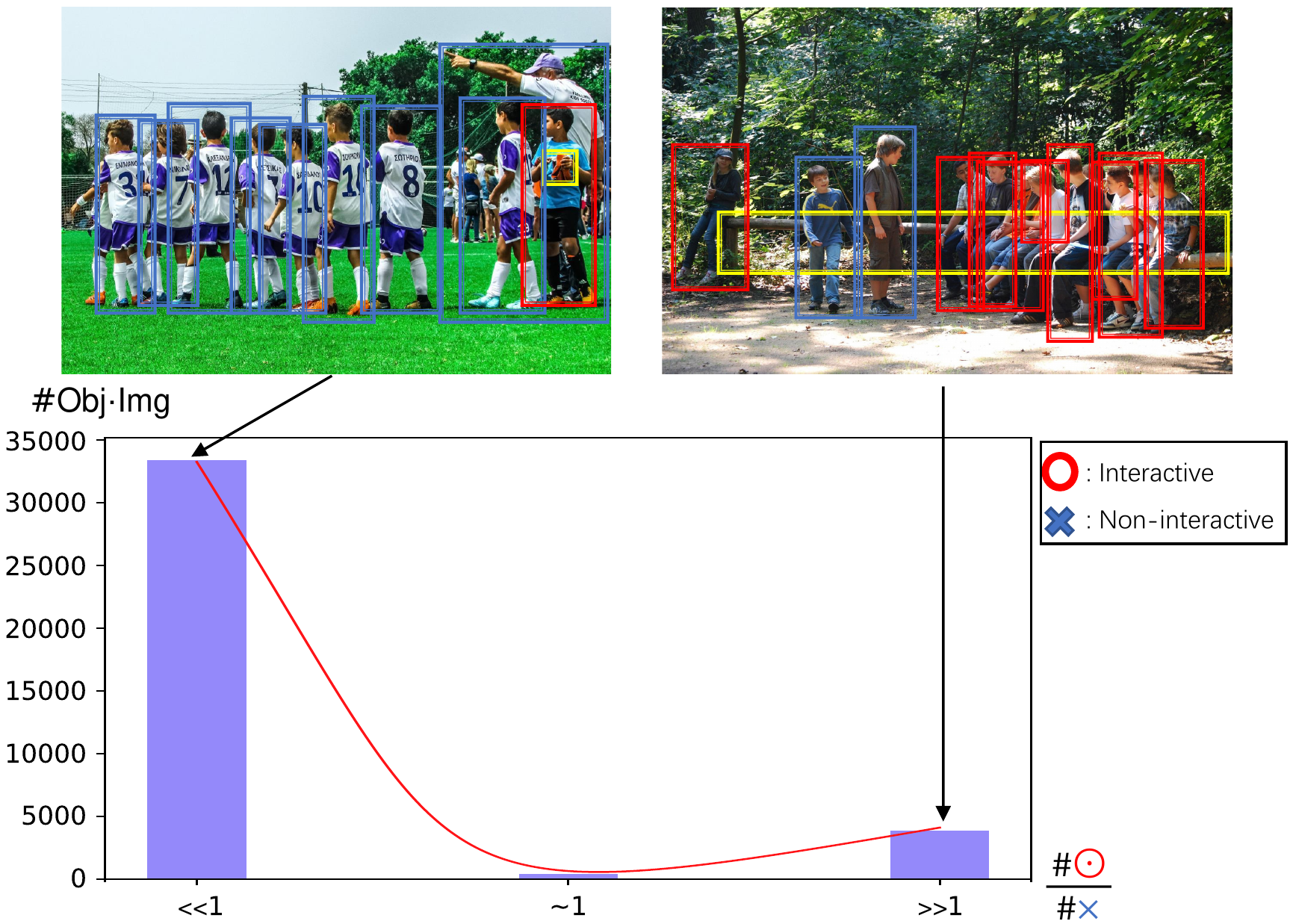}
	\vspace{-10px}
	\caption{Distribution of interactive ratio between interactive and non-interactive H-O pairs in HICO-DET~\cite{hicodet}, where two representative samples are shown. For the pairs containing a given object (yellow), either non-interactive pairs or interactive pairs dominates, with the former much more frequent.
	}
  \label{fig:insight}
  \vspace{-18px}
\end{figure}

Overall, HOI detection can be divided into \textit{H/O localization}, \textit{interactive H-O pairing}, i.e., localizing the interactive humans and objects and pairing them correctly, and \textit{verb classification}. 
The most conventional approach is the two-stage paradigm~\cite{interactiveness,gao2018ican,pmfnet,NoFrills,djrn} proposed in HO-RCNN~\cite{hico}, where an object detector is first adopted to detect all the human/object instances in a given image, followed by \textit{exhaustive pairing} and verb classification. The major issue of this straightforward approach 
is that, in practice, only a small portion of human/object instances are involved in HOI relationships, making the exhaustive object detection and pairing excessive and seemingly unnecessary. 

The other approach consists of one-stage methods~\cite{hotr,qpic} represented by PPDM~\cite{ppdm}. One-stage approach adopts an \textit{end-to-end} manner following the one-stage object detection~\cite{centernet,cornernet}, where the object boxes are replaced by H-O pair boxes and the object category by HOI category. This circumvents the exhaustive instance detection and explicit pairing while achieving the same goal. However, given that a typical image, e.g., HICO-DET~\cite{hicodet} contains 2.47 HOIs on average, it is still unsatisfactory that a recent state-of-the-art one-stage method QPIC~\cite{qpic} still needs \textbf{100} output pairs per image to achieve a recall of 70\%. 

Though significant progress has been made, the two paradigms are still bottle-necked by \textbf{H-O pairing}: they fail to effectively extract interactive pairs but generate excessively redundant and negative pairs. 
One of the early studies to address this problem is TIN~\cite{interactiveness,li2021transferable}, where the pairing problem is addressed by interactiveness learning. A \textit{pair-wise} interactiveness binary classifier is inserted to discriminate whether a human and an object should be paired (i.e., interactive or otherwise). Despite its simple design, the improvement is rather decent, indicating the great potential of such proper pairing strategies.

Given this early promise, here, we aim at improving HOI detection by studying the interactiveness problem from a \textbf{global} and \textbf{distribution} point of view. 
Specifically, we propose a previously overlooked but powerful prior: the \textbf{\extrema} property of interactiveness. 
In Figure~\ref{fig:insight}, the dominating proportion of H-O pairs given the same object in an image are either interactive or non-interactive, while most of the time they are non-interactive. 
This phenomenon of interactiveness distribution is closely related to Zipf's Law~\cite{auerbach1913gesetz}: \textit{informative events are rarer than non-informative events}. 
To exploit this prior, we pursue a verb-agnostic measurement of interactiveness.
In line with the notion of field and its global measurement~\cite{feynman1965feynman} as such, we introduce the \textbf{``interactiveness field''} to model the global interactiveness distribution of HOI images. 
Specifically, we encode the H-O pairs in a complex scene as a field.
Each pair is encoded as a point with an ``energy'' value, indicating its difference from other pairs. 
The field is expected to obey the \extrema prior, \ie, the high-energy pairs should be rare.
Based on this, we analyze the change of the field with the modification on a {\em single} pair and impose energy constraints on the field modeling: modification on high-energy pairs should bring more salient influence.
Then, the interactiveness labels are bounded with the modeled field following the prior.

To use the interactiveness field, we propose a novel paradigm. 
First, instead of exhaustive human/object detection, a DETR~\cite{detr} structure detector is adopted to directly detect initial H-O pairs organized in an object-centric manner.
Subsequently, based on the interactiveness field subjecting to the \extrema prior, we design an interactiveness field module to further filter out non-interactive pairs. 
Finally, the filtered pairs are fed into a verb classifier for HOI classification.
On HICO-DET~\cite{hicodet} and V-COCO~\cite{vcoco}, we achieve state-of-the-art and significant improvements. 

Our contribution includes: 
\textbf{1)} the interactiveness \extrema prior of HOI is identified as a key to improve the H-O pair filtering and boost the HOI detection, based on which an interactiveness field model is introduced;
\textbf{2)} we achieve state-of-the-art performance on widely-used HOI benchmarks.

\section{Related Works}
\label{sec:related}
Rapid progress has recently been made in HOI learning.
Many large datasets~\cite{hicodet,vcoco,OpenImages,pastanet} and deep learning based methods~\cite{Gkioxari2017Detecting,gao2018ican,interactiveness,gpnn,NoFrills,pmfnet,analogy,pastanet,djrn,DRG,vcl,idn,hou2021atl,kim2020detecting,qpic} have been proposed. For example, Chao~\etal~\cite{hicodet} proposed the widely-used multi-stream framework, while GPNN~\cite{gpnn} and Wang~\etal~\cite{wang2020contextual} adopted graphs to model the HOI relationship. 
iCAN~\cite{gao2018ican} and PMFNet~\cite{pmfnet} adopted the self-attention mechanism to correlate the human, object, and context from different levels. TIN~\cite{interactiveness} introduced interactiveness to filter out non-interactive pairs.
Besides, some works~\cite{analogy,kim2020detecting,zhong2020polysemy} focused on the relationship between HOIs.
In terms of information utilization, DJ-RN~\cite{djrn} introduced 3D information for better inference. PaStaNet~\cite{pastanet} introduced part states as an intermediate semantic hierarchy for further HOI reasoning. DRG~\cite{DRG} considered HOI from both human-centric and object-centric point of view, while VCL~\cite{vcl} exploited the compositional characteristic of HOI. IDN~\cite{idn} analyzed how HOI is integrated and composed from a transformation-based perspective.

Recently, several one-stage methods have been proposed~\cite{ppdm,uniondet,ipnet,qpic}, where parallel HOI detectors directly detect HOIs triplets, in contrast to the conventional two-stage method~\cite{gao2018ican,interactiveness} for interaction prediction. 
PPDM~\cite{ppdm}, UnionDet~\cite{uniondet}, and IP-Net~\cite{ipnet} adopted a variant of one-stage object detector~\cite{centernet,cornernet} for HOI detection. 

While based on the recently proposed transformer detector DETR~\cite{detr}, QPIC~\cite{qpic} managed to achieve impressive performance. By capitalizing on the powerful transformer, DETR~\cite{detr} achieved impressive performance without many hand-designed components. A fixed-size set of predictions is produced in a single pass through the decoder. The main loss is calculated by matching the predicted and ground-truth predictions via an optimal bipartite matching, followed by imposing the specific losses. QPIC~\cite{qpic} adapted the paradigm by regressing both the human and object box with the addition of a verb classifier to detect HOI triplets.

\section{Methods}
Our goal is to address the pairing problem in HOI detection, by exploiting the underlying distributional information of H-O pairs subject to the interactiveness \extrema prior. 
Section~\ref{sec:prelim} first presents the preliminaries of our method and a formal definition of interactiveness field.
Then, in Section~\ref{sec:pdm}, we introduce how interactiveness field is modeled with the pair distributional characteristics. 
In Section~\ref{sec:sys}, we  demonstrate how to design the practical system. 

\subsection{Preliminaries}
\label{sec:prelim}
\begin{figure}
    \centering
    \includegraphics[width=0.36\textwidth]{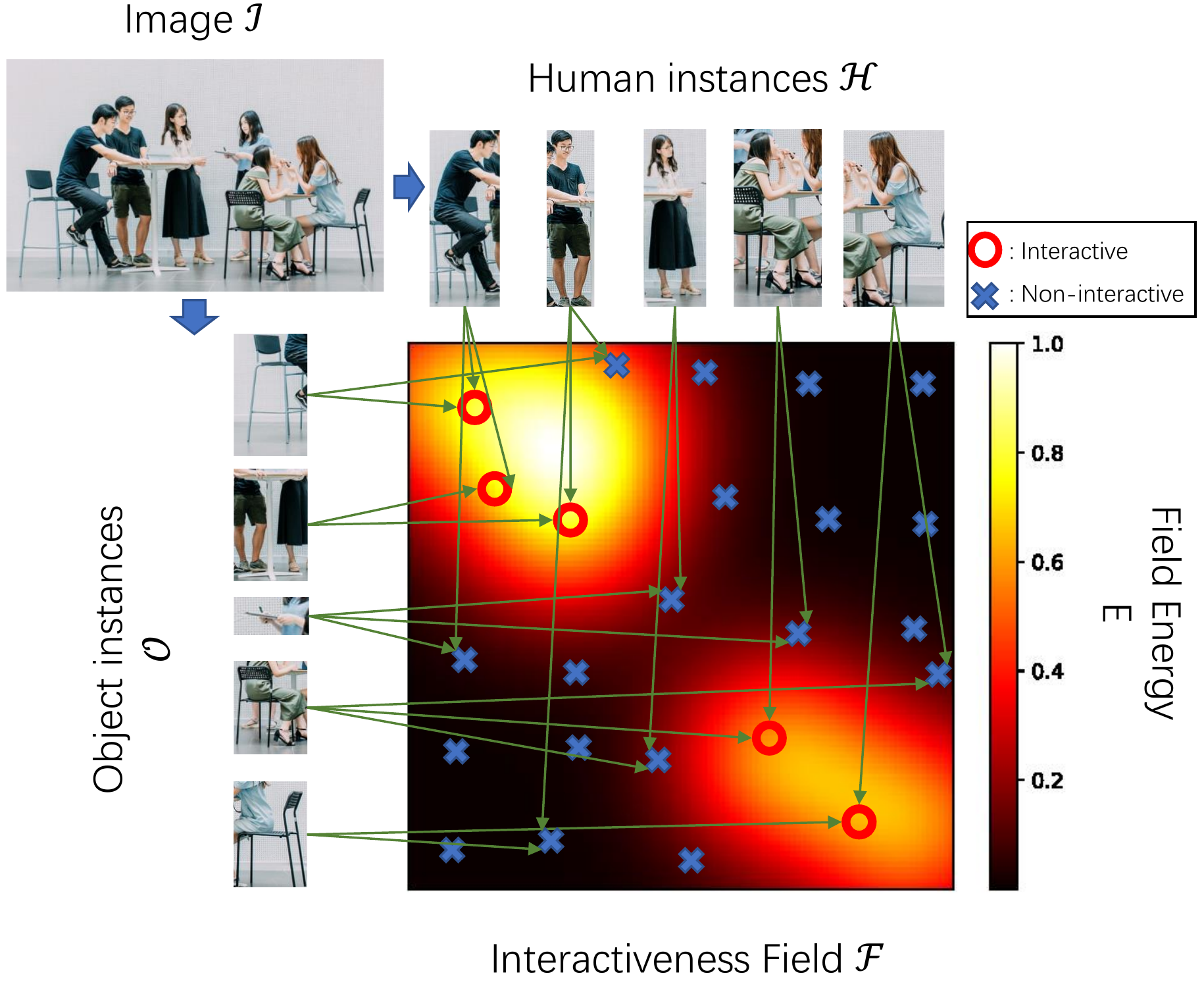}
    \vspace{-10px}
    \caption{Interactiveness field illustration.}
    \label{fig:field}
    \vspace{-15px}
\end{figure}

Given an image $\mathcal{I}$, we define \textit{interactiveness field} $\mathcal{F}$ as 
\begin{equation}
    \label{eq:def-general}
    \mathcal{F}=(\mathcal{A}\times \mathcal{A}, E(\cdot):\mathcal{A}\times \mathcal{A}\to [0, 1]),
\end{equation}
where $\mathcal{A}$ denotes arbitrary areas in $\mathcal{I}$, $E(\cdot)$ is the \textit{energy} function for each area pair, indicating the \textit{relative difference} of each pair against other pairs. 
Given the interactiveness \extrema prior, the energy function is closely related to the interactiveness: \textit{when the pairs are mostly non-interactive, interactive pairs would possess high energy and vice versa}.

Since we focus on HOI detection, where only human/object instances are considered to be potentially interactive, the definition in Eq.~\ref{eq:def-general} is simplified as 
\begin{equation}
    \mathcal{F}=(\mathcal{P}=\mathcal{H}\times \mathcal{O}, E(\cdot):\mathcal{P}\to [0, 1]),
\end{equation}
where $\mathcal{H},\mathcal{O}$ are the human and object instance proposals in $\mathcal{I}$ respectively, as illustrated in Figure~\ref{fig:field}.

Here, we focus on the pairs concerning the same given object $o_i$.
Each pair $\langle h_i,o_i \rangle$ is represented by the extracted feature $f_{\mathcal{P}}^i\in f_{\mathcal{P}}$, and $E(\cdot)$ is implemented by specially designed neural networks.
Thus, the interactiveness field $\mathcal{F}$ could be generally formulated as
\begin{equation}
    \label{eq:field}
        \mathcal{F} = (f_{\mathcal{P}}, E(\cdot)),\ f_s = g(f_{\mathcal{P}}), \\
\end{equation}
where $f_s$ denotes the \textbf{summary} of the field extracted from the pairs with summary function $g(\cdot)$, the energy function $E(\cdot)$ takes the sample feature $f_{\mathcal{P}}^i$, producing the energy of the input sample.
Intuitively, the binary pair-wise classifier introduced in TIN~\cite{interactiveness} could be a simple implementation of $E(\cdot)$, lacking the consideration of global interactiveness distribution and pair difference. 
However, in Section~\ref{sec:experiment}, we show that without the interactiveness \extrema prior, the simple TIN-style classifier outputs a biased interactiveness score thus performs unsatisfactorily on interactiveness discrimination.
That is, for almost all the pairs in an image involving the same object, near-zero interactiveness score is produced due to the extreme imbalance in data distribution.
Rather than resorting to simple modeling using a pair-wise classifier, we propose to model the interactiveness field regulated by the interactiveness \extrema prior, considering the underlying global-distribution properties.

\subsection{Interactiveness Field Modeling}
\label{sec:pdm}

In the following, we first delve into how interactiveness field is modeled in Section~\ref{sec:rarity2field} subject to the interactiveness \extrema prior. Notably,
two main constraints are derived in Section~\ref{sec:change2field} to regulate the field, where the global change in $\mathcal{F}$ upon removing or modifying a single local pair will be analyzed.
The modeling formulation detailed in Sections~\ref{sec:rarity2field}--\ref{sec:change2field}  only requires the interactiveness \extrema prior. In Section~\ref{sec:field2bin}, we  describe how the interactiveness labels can then be incorporated into the formulation to enhance the proposed field modeling.

\subsubsection{Cardinality Constraint}
\label{sec:rarity2field}
\begin{figure}[!t]
    \centering
    \includegraphics[width=0.4\textwidth]{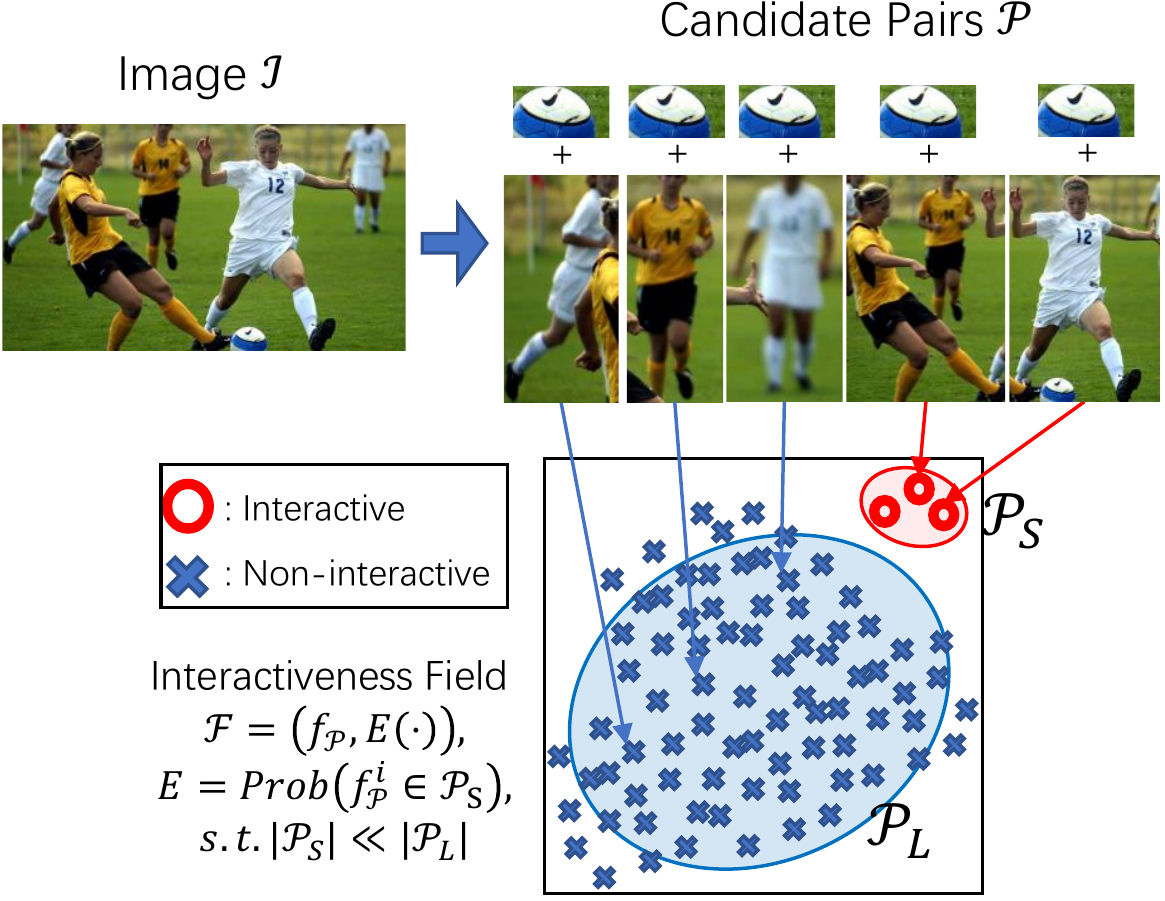}
    \vspace{-10px}
    \caption{Interactiveness field modeling subject to the interactiveness \extrema prior.}
    \label{fig:field_modeling}
    \vspace{-20px}
\end{figure}

As illustrated in Figure~\ref{fig:insight}, candidate pairs involving the same object can be divided into two clusters: the \textit{rare, high-energy} cluster and the \textit{frequent, low-energy} cluster. 
Correspondingly, we argue that the interactiveness field should possess the following property: candidate pairs set $\mathcal{P}$ should consist of  two diverse sets $\mathcal{P}_S$ and $\mathcal{P}_L$ with salient differences in cardinality. This property is formulated as 
\begin{equation}
    \begin{aligned}
    &\mathcal{P} = \mathcal{P}_L \cup \mathcal{P}_S, \\
    & \text{s.t.}\ \mathcal{P}_L \cap \mathcal{P}_S = \O,\ |\mathcal{P}_S| \ll |\mathcal{P}_L|,
    \end{aligned}
\end{equation}
where $|\cdot|$ denotes cardinality.
The interactiveness field is 
\begin{equation}
    \label{eq:field_dnn}
    \begin{aligned}
    &\mathcal{F} = (f_{\mathcal{P}}, E(\cdot)),\ E(f_{\mathcal{P}}^i)=\mathit{Prob}(\mathcal{P}^i \in \mathcal{P}_S),\\
    & \text{s.t.}\ |\mathcal{P}_S| \ll |\mathcal{P}_L|.
    \end{aligned}
\end{equation}

Thus, given the extracted pair feature $f_\mathcal{P} \in \mathcal{R}^{N\times C}$, the summary function $g(\cdot)$ first extracts the two clusters $\mathcal{P}_S$ and $\mathcal{P}_L$, denoted by centroids $c_s, c_l \in \mathcal{R}^C$ and assignment vectors $A_s, A_l\in \mathcal{R}^N$, where $A_s^i, A_l^i$ respectively mean the probability that pair $i$ belongs to cluster $\mathcal{P}_S,\mathcal{P}_L$, subjecting to $\sum_iA_s^i \ll \sum_iA_l^i$.
$f_s=(c_s, c_l)$ is then adopted as the summary representation of the interactiveness field $\mathcal{F}$. 
The energy function $E(\mathcal{P}^i) = A_s^i$ for each pair $\mathcal{P}^i$ is given by the probability that the pair belongs to $\mathcal{P}_S$. 
Figure~\ref{fig:field_modeling} illustrates the formulation:
\begin{equation}
    \begin{aligned}
    c_s, c_l, A_s, A_l &= g(f_\mathcal{P}), \text{s.t.}\ \sum_iA_s^i \ll \sum_iA_l^i.
    \end{aligned}
\end{equation}

To regulate the field to satisfy the interactiveness \extrema prior, a cardinality loss $L_{\text{card}}$ is formulated as 
\begin{equation}
    L_{\text{card}} = \sum_iA_s^i - \sum_iA_l^i.
\end{equation}
The loss corresponds to the constraint $\sum_iA_s^i \ll \sum_iA_l^i$, which encourages more pronounced cardinality difference.
Noticeably, here we do not need the binary interactiveness labels~\cite{interactiveness} in modeling. 
Thus, the above modeling can be regarded as an \textbf{unsupervised} process using our \extrema prior. 
In Section~\ref{sec:field2bin}, we introduce how to further enhance the interactiveness discrimination with the binary labels. 

\subsubsection{Field Change Constraints}
\label{sec:change2field}The cardinality constraint introduced above focuses
on the \textit{static} status of the interactiveness field. 
We now investigate how to model the field by observing how $\mathcal{F}$ should change upon modifying local pairs with different energy level.

\noindent{\bf Field Change against Pair Removal.}
\begin{figure}[!t]
    \centering
    \includegraphics[width=0.4\textwidth]{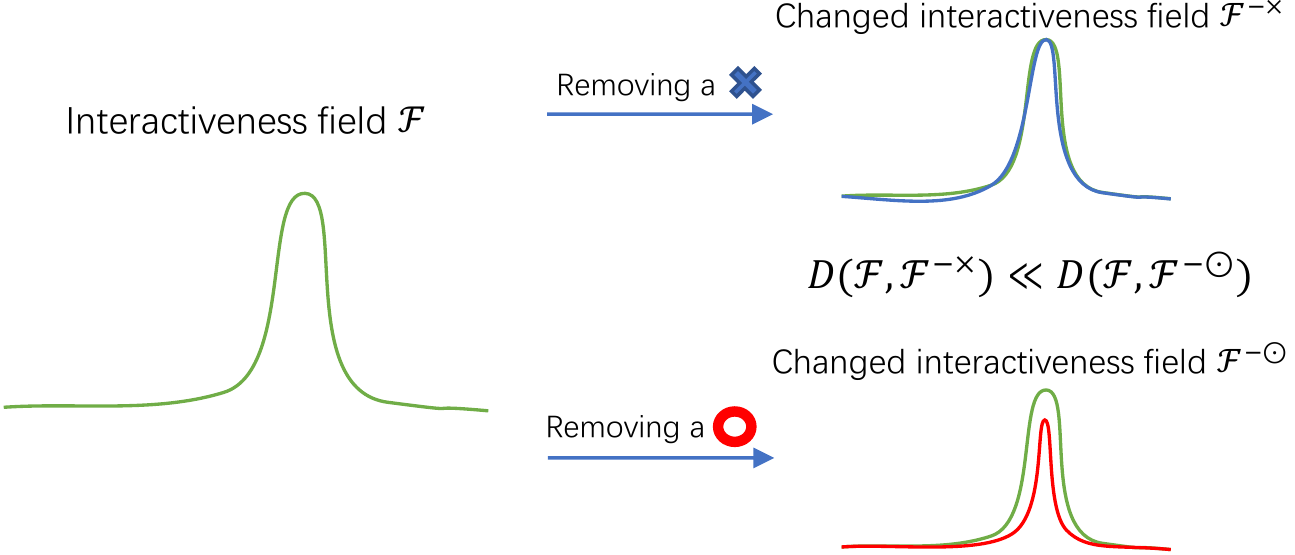}
    \vspace{-10px}
    \caption{Field change against pair removal. Removal rare pairs (usually also interactive) brings more salient change.}
    \label{fig:shift2bin}
    \vspace{-15px}
\end{figure}
We first explore how the global field representation changes when a certain sample is removed.
Starting from the interactiveness field $\mathcal{F}$ in Section~\ref{sec:rarity2field}, we can tell the removal of a high-energy point would affect the overall representation of $\mathcal{F}$ more than the removal of a low-energy point (Figure~\ref{fig:shift2bin}).
So we adopt a difference indicator $D_r$ to encode the global field change when a certain sample is removed, which is formulated as
\begin{equation}
    \begin{aligned}
        D_r^i &= D(\mathcal{F}, \mathcal{F}^{-i}), \\
        \mathcal{F} &= (f_{\mathcal{P}}, E(\cdot)),\ \mathcal{F}^{-i} = (f_{\mathcal{P}}^{-i}, E(\cdot)),
    \end{aligned}
\end{equation}
where $D(\cdot, \cdot)$ denotes the difference between the two fields, and $f_{\mathcal{P}}^{-i}=f_{\mathcal{P}} / f_{\mathcal{P}}^i$ denotes the pair features minus $f_{\mathcal{P}}^i$.

Based on this, given the pair feature $f_\mathcal{P}$, $g(\cdot)$ (defined in Section~\ref{sec:rarity2field}) first extracts the field summary representation $f_s=(c_s, c_l)$ for $\mathcal{F}$.
Then, each pair $i$ is removed, and the rest pair features $f_{\mathcal{P}}^{-i}$ are fed to $g(\cdot)$, which produces the modified field representation $f_s^{-i}$.
The L2 distance between $f_s$ and $f_s^{-i}$ is then defined as the difference indicator $D_r^i$.
Larger $D_r^i$ indicates that the pertinent pair is more likely to  have higher energy level (or more different from the other pairs).
The above process can be summarized as 
\begin{equation}
    \begin{aligned}
    f_s &= (c_s, c_l) = g(f_\mathcal{P}), \\
    f_s^{-i} &= (c_s^{-i}, c_l^{-i}) = g(f_\mathcal{P}^{-i}), \\
    D_r^i &= \|f_s, f_s^{-i}\|_2.
    \end{aligned}
\end{equation}

Since the removal of a pair will definitely change the field, instead of enforcing $D_r$ to be zero for frequent low-energy pairs, a rank loss $L_{\text{rank}}^r$ is imposed as 
\begin{equation}
    \label{eq:rank_loss}
    \begin{aligned}
    L_{\text{rank}}^r &= \sum_{i \in \mathcal{P}_S}\sum_{j \in \mathcal{P}_L}D_r^j - D_r^i, \\
    \mathcal{P}_S &= \{i:A_s^i > A_l^i\}, \ \mathcal{P}_L = \{i:A_l^i > A_s^i\},
    \end{aligned}
\end{equation}
where $A_l, A_s$ are the assignment vectors produced by $g(f_{\mathcal{P}})$.
$L_{\text{rank}}^r$ only encourages the assumed high-energy pairs to cause more field change with their removal than the low-energy pairs.

\noindent{\bf Field Change against Pair Modification.}
\begin{figure}[!t]
    \centering
    \includegraphics[width=0.4\textwidth]{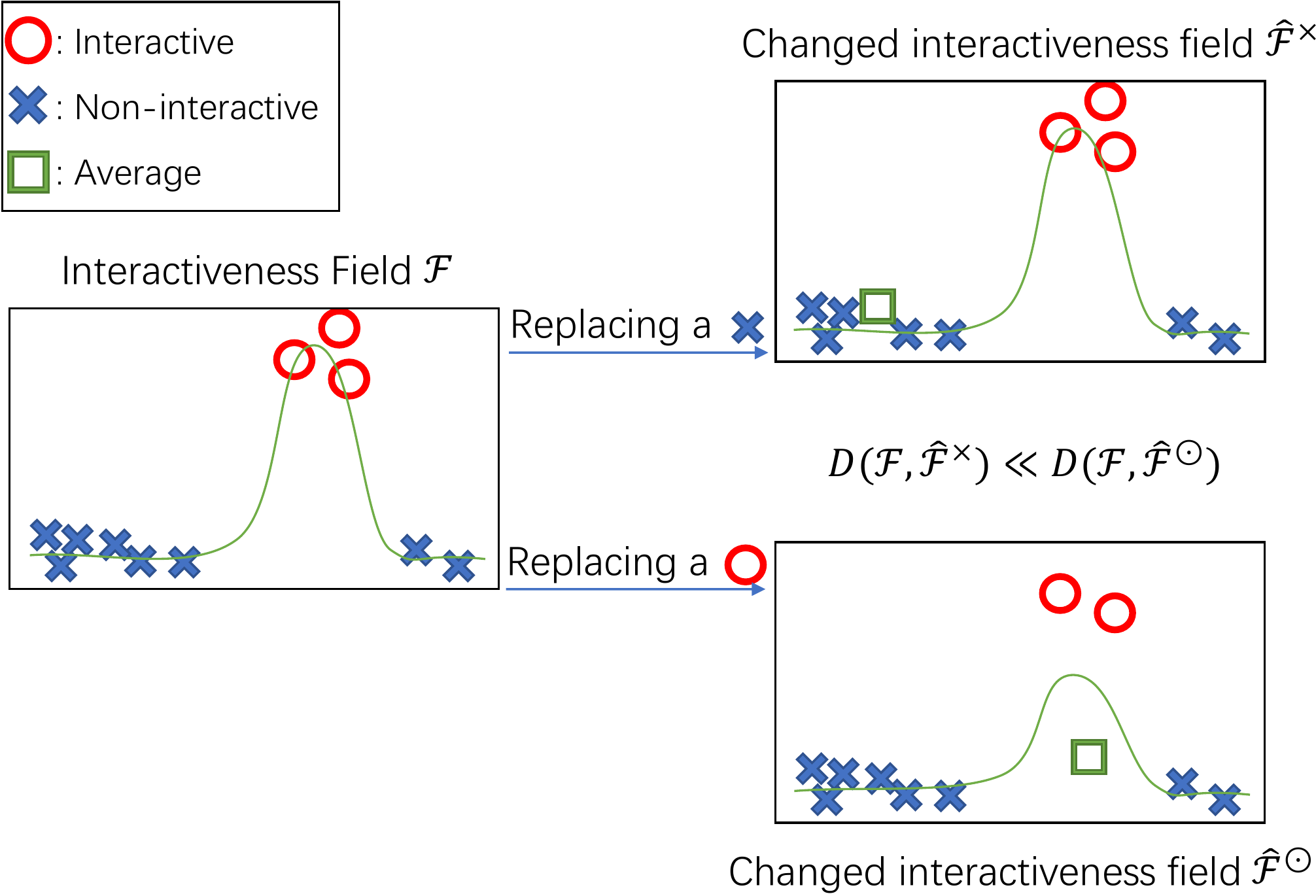}
    \vspace{-10px}
    \caption{Field change against pair modification. Modification on rare pairs (usually also interactive) brings more change.}
    \label{fig:flow2bin}
    \vspace{-15px}
\end{figure}
Another worthwhile constraint to explore is how the field changes when a pair is modified, in our case, replaced by the \textbf{mean} pair representation.
Still referring  to the interactiveness field $\mathcal{F}$ in Section~\ref{sec:rarity2field}, given  a field with most areas possessing low energy, we could tell that the mean representation of this field should also carry low energy.
Thus, if we replace a high-energy pair with the mean, the overall field representation should change significantly.
On the other hand, the overall field representation should not change much when a low-energy pair is replaced by the mean.
Thus, we can obtain another difference indicator $D_m$ as 
\begin{equation}
    \label{eq:flow2bin}
    \begin{aligned}
        D_m^i &= D(\mathcal{F}, \hat{\mathcal{F}}^i), \\
        \mathcal{F} &= (f_{\mathcal{P}}, E(\cdot)),\ \hat{\mathcal{F}}^i = (\hat{f}_{\mathcal{P}}^i, E(\cdot)).
    \end{aligned}
\end{equation}
$\hat{f}_{\mathcal{P}}^i$ denotes $f_{\mathcal{P}}$ replacing $f_{\mathcal{P}}^i$ with mean representation $\bar{f}_{\mathcal{P}}$.

To implement the above, the field representation $f_s$ is first extracted by $g(\cdot)$ in Section~\ref{sec:rarity2field}. 
Then we obtain the modified field $\hat{f}_s^i$ by feeding $\hat{f}_\mathcal{P}^i$ to $g(\cdot)$. 
The difference between $f_s$ and $\hat{f}_s^i$ is defined as the difference indicator $D_m^i=\|f_s - \hat{f}_s^i\|$.
Again, larger difference indicates the sample is more likely to be a high-energy pair.
The rank loss $L_{\text{rank}}^m$ with the same formulation as Eq.~\ref{eq:rank_loss} is computed.

\subsubsection{Binding with Interactiveness Labels}
\label{sec:field2bin}
The previous modeling formulation only adopts the interactiveness \extrema prior, functioning in an unsupervised manner. 
For further enhancement, we can  bind the field with the interactive semantics via specially designed losses to connect interactiveness labels transferred from HOI labels, following TIN~\cite{interactiveness}. This  encourages the modeled field to simultaneously approach the ground truth distribution while following the prior when applicable.

Following the set-based training procedure in QPIC~\cite{qpic}, the interactiveness labels are assigned to the candidate pairs.
Given the assigned labels, we obtain the correspondence between $\{\mathcal{P}_S,\mathcal{P}_L\}$ and \{interactive pairs, non-interactive pairs\}.
In the following, we assume $\mathcal{P}_S$ is  interactive for ease of description, which is most of the cases. Analogous descriptions apply when $\mathcal{P}_L$ is interactive.
A simple cross entropy loss $L_{\text{ce}}$ is imposed on $A_s, A_l$.
Then, the cardinality loss in Section~\ref{sec:rarity2field} is enriched with an additional term:
\begin{equation}
    L_{\text{card}} = \sum_iA_s^i - \sum_iA_l^i + \| n_T - \sum_iA_s^i\|,
\end{equation}
where $n_T$ is the number of interactive pairs for this object in this image.
This added term regulates the cardinality of $\mathcal{P}_S$ to be the same as the number of interactive pairs. 
Moreover, a clustering loss $L_{\text{clus}}$ inspired by \cite{rebuffi2021lsd} is formulated as 
\begin{equation}
    \begin{aligned}
    &p_{ij}=A_s^iA_s^j + A_l^iA_l^j, \\
    &L_{\text{clus}}=\sum_{i,j}\left((\alpha_{ij} - 1)\log(1 - p_{ij}) - \alpha_{ij}\log p_{ij}\right),
    \end{aligned}
\end{equation}
where $\alpha_{ij}=1$ if pair $i,j$ are both interactive or non-interactive, otherwise $\alpha_{ij}=0$. 
This loss encourages pairs with the same interactiveness label to be clustered together.

With these losses, we force the field  $\mathcal{F}$ to simultaneously follow the interactiveness \extrema prior while approaching the ground-truth interactiveness distribution. 
More discussions on the generalization of our interactiveness bi-modal prior would be included in the appendices.

\subsection{Practical System Design}
\label{sec:sys}
\begin{figure*}[!t]
    \centering
    \includegraphics[width=0.8\textwidth]{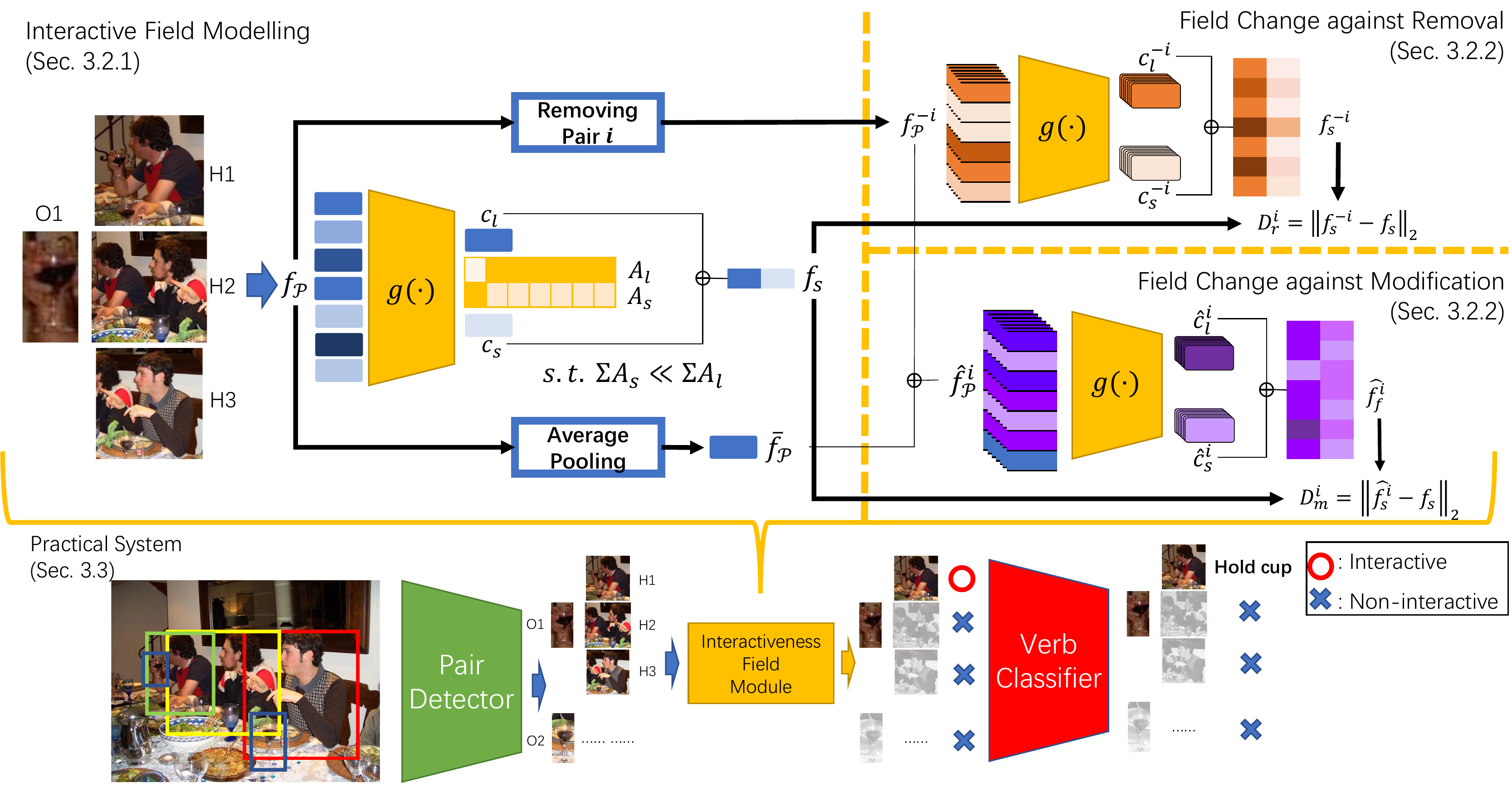}
    \vspace{-10px}
    \caption{Our pipeline for HOI detection with interactiveness field modeling which is composed of four components. Visual feature extractor generates visual feature map $f$, based on which pair decoder decodes the candidate pairs $\mathcal{P}$ along with feature $f_\mathcal{P}$. Our proposed interactiveness field module models the interactiveness field and assigns interactiveness score $S_b$ for each pair. The verb decoder infers the verb score $S_v$ for generating the final score as $S=S_v\cdot S_b$.}
    \label{fig:pipeline}
    \vspace{-19px}
\end{figure*}

Next, we introduce how the interactive field is incorporated into a practical HOI detection system. 
Such system contains four components: visual feature extractor, pair decoder, interactiveness field module defined in Section~\ref{sec:pdm}, and verb classifier.
Figure~\ref{fig:pipeline} shows the overall pipeline.

\subsubsection{Visual Feature Extractor}
Our feature extractor is a combination of a CNN and a transformer encoder. 
In detail, given an image $\mathcal{I} \in \mathcal{R}^{H\times W \times 3}$, the CNN encodes it into feature map $f_C \in \mathcal{R}^{H'\times W' \times C_C}$, which is linearly projected to a lower dimension of $C_T$, flattened into $\mathcal{R}^{(H'W') \times C_T}$, which is then fed into the transformer encoder with sinusoidal positional embedding $E \in \mathcal{R}^{(H'W') \times C_T}$ to output the final visual feature $f \in \mathcal{R}^{(H'W') \times C_T}$. 
The CNN encoder aggregates the local information into patch tokens, while the transformer encoder, leveraging the power of multi-head self-attention, generates a feature map with rich global contextual information. 

\subsubsection{Pair Decoder}
A transformer decoder is adopted as the pair decoder. 
With visual feature $f$ as $K,V$, a learned query embedding $Q \in \mathcal{R}^{M \times C_T}$ is utilized to decode the candidate pairs $\mathcal{P}$ along with feature $f_\mathcal{P}$. 
A fully-connected layer is imposed on $f_\mathcal{P}$ to classify the corresponding object class $o$, and two two-layer MLPs regress the human and object box coordinates $b^h, b^o$.
Following previous set-based training process~\cite{detr, qpic}, with the Hungarian bipartite matching algorithm, ground truth labels are assigned to the pair predictions. 
Multiple loss items are computed, including generalized IoU (Intersection over Union) loss $L_{\text{giou}}^h, L_{\text{giou}}^o$, box regression L1 loss $L_{\text{reg}}^h, L_{\text{reg}}^o$, and object class cross-energy loss $L_{o}$. 
The pair decoder is first trained along with the visual feature extractor with target loss
\begin{equation}
    L_{\text{pair}} = \lambda_1(L_{\text{giou}}^h + L_{\text{giou}}^o) + \lambda_2(L_{\text{reg}}^h + L_{\text{reg}}^o) +\\ \lambda_3L_{o},
\end{equation}
where $\lambda_1, \lambda_2, \lambda_3$ are weighting coefficients.

\subsubsection{Implementation of Interactiveness Field Module}

To implement the interactiveness field module, multiple choices for $E(\cdot)$ and $g(\cdot)$ are proposed.
A toy design is first used, where
$E(\cdot)$ and $g(\cdot)$ are implemented as a hierarchical cluster followed by a \textit{soft} two-means cluster with the hierarchical centroids as the initial centroids. 
For both clustering procedures, Euclidean distance is adopted. 
By ``soft'', we mean the distance vectors $D_s, D_l\in \mathcal{R}^N$ are respectively processed by a softmax function along each column to obtain the assignment vectors $A_s, A_l$.

For a more advanced version, the two-means clustering is replaced by a modified multi-head attention layer. 
In detail, it takes $f_\mathcal{P}$ as $K,V$, and the two hierarchical cluster centroids as $Q$ to extract $C$. To obtain the assignment matrix, the original softmax function used to generate attention from logits is replaced by sigmoid function following averaging, where the attention value before averaging is adopted as assignment matrix $A$.
In this way, the multi-head attention module is adapted for clustering by regarding the attention mechanism as a soft assignment procedure, thus acquiring a more powerful mean field representation.
The target loss is formulated as 
\begin{equation}
\vspace{-0.05in}
    L_{\text{field}} = \lambda_4L_{\text{card}} + \lambda_5L_{\text{ce}} +\\ \lambda_6L_{\text{clus}} + \lambda_{r}(L_{\text{rank}}^r + L_{\text{rank}}^m),
    \vspace{-0.05in}
\end{equation}
where $\lambda_4, \lambda_5, \lambda_6, \lambda_{r}$ are weighting coefficients, and the different loss terms have already been defined in Section~\ref{sec:pdm}.

\vspace{-5px}
\subsubsection{Verb Decoder}
Another transformer decoder takes $f$ (whole image feature) as $K,V$, $f_\mathcal{P}$ as $Q$, followed by a fully-connected verb classifier, which is used to produce the verb score $S_v$. 
The verb classifier is attached with verb label cross-energy loss $L_{\text{verb}}$. 

\subsubsection{Training and Inference on HOI Datasets}
The training is divided into three stages.
First, we train the pair decoder along with the visual feature extractor using $L_{\text{pair}}$.
Then, the interactiveness field module is introduced and the three components are fine-tuned together with loss $L=L_{\text{pair}} + L_{\text{field}}$.
Finally, the verb classifier is included, and the whole system is trained with $L=L_{\text{pair}} + L_{\text{field}} + L_{\text{verb}}$.

In some cases interactive pairs dominate, e.g., in a restaurant, several humans are \textit{sitting beside} the dinner table except for the waiter. 
We consider this special situation in training. 
Since these cases only account for less than 10\% in HICO-DET~\cite{hicodet}, we assume that the interactive pairs are always  minorities in inference.
Thus, the energy and difference indicators can be directly adopted to compute interactiveness binary score $S_b$.
The difference indicators are aggregated and normalized to $[0,1]$, and then combined with $A_s$, producing $S_b=(A_s+(\sigma(D_r)+\sigma(D_m) - 1)) / 2 \in [0, 1]$, where $\sigma(\cdot)$ is sigmoid function. 
The final prediction is constructed as $(b^h, b^o, o, S) \in \mathcal{P}_r$, where $S=S_v \cdot S_b$.
Our experimental results show even with this compromised strategy, the improvement is still substantial. 

Notwithstanding, a possible problem is that though the interactiveness \extrema prior is statistically reasonable, there still exist exceptions, e.g., an image contains only one person.
For the practical system here, we cover the situation with sparse scene in two ways.
First, the human proposals $\mathcal{H}$ generated by the model are abundant most of the time, making the prior still applicable.
Second, pairs with the same object category are aggregated and modeled by the same field, as they share similar interactiveness patterns.

\section{Experiments}
\label{sec:experiment}

\subsection{Dataset and Metric}
\label{sec:dataset}
We adopt two large-scale HOI detection benchmarks: HICO-DET~\cite{hicodet} and V-COCO~\cite{vcoco} for evaluation.
{\bf HICO-DET~\cite{hicodet}} consists of 38,118 training images, 9,658 testing images, 600 HOI categories (comprising of 80 COCO~\cite{coco} objects and 117 verbs), and more than 150 K annotated HOI pairs. We use
mAP for evaluation: true positive is required to contain accurate human and object locations (box IoU with reference to GT box is larger than 0.5) and accurate interaction classification. 
Following \cite{hicodet}, mAP for three sets: Full (600 HOIs), Rare (138 HOIs), Non-Rare (462 HOIs) under both Default and Known Object modes are reported. 
{\bf V-COCO~\cite{vcoco}} contains 10,346 images (2,533 in train set, 2,867 in validation set, and 4,946 in test set), and covers 29 verb categories (25 HOIs and 4 body motions) and 80 objects from COCO~\cite{coco}.
We use role mean average precision under both scenario 1 and scenario 2 as evaluation metrics, where only the 25 HOIs are taken into consideration.

\subsection{Implementation Details} 
\label{sec:implementation}

We adopt ResNet-50 followed by a six-layer transformer encoder as our visual feature extractor. 
The pair decoder and the verb decoder are both implemented as a six-layer transformer decoder.
During training, AdamW~\cite{loshchilov2017decoupled} with the weight decay of 1e-4 is used.
The visual feature extractor and pair decoder are initialized from COCO~\cite{detr} pre-trained DETR~\cite{detr}. 
The query size is set as 64 for HICO-DET~\cite{hicodet} and 100 for V-COCO~\cite{vcoco} following CDN~\cite{cdn}.
The loss weight coefficients $\lambda_1, \lambda_2, \lambda_3$ are respectively set as 1, 2.5, 1, exactly the same as QPIC~\cite{qpic}. 
The visual feature extractor and pair decoder are fine-tuned for 90 epochs with a learning rate of 1e-4 which is decreased by 10 times at the 60th epoch.
Then, the interactiveness field module is introduced and fine-tuned for another 9 epochs with learning rate of 1e-4.
Finally, the verb decoder is added and the whole model is trained for 30 epochs.
All experiments are conducted on four NVIDIA GeForce RTX 3090 GPUs with batch size of 16.
In inference, a pair-wise NMS with threshold of 0.6 is conducted. That is, low-score predictions with both human and object $\text{IoU}>0.6$ compared to the same category high-score pair is suppressed.

\subsection{Results}
\label{sec:res}

\noindent{\bf Results on HOI Detection Benchmarks}
We first report the results on HICO-DET~\cite{hicodet}.
\begin{table}[t]
    \centering
    \resizebox{0.8\linewidth}{!}{\setlength{\tabcolsep}{0.8mm}{
    \begin{tabular}{l c c c c c c }
        \hline
                                                 & \multicolumn{3}{c}{mAP Default $\uparrow$} &\multicolumn{3}{c}{mAP Known Object $\uparrow$} \\
        Method                                   & Full & Rare & Non-Rare & Full & Rare & Non-Rare \\
        \hline
        \hline
        iCAN~\cite{gao2018ican}                  & 14.84 & 10.45 & 16.15 & 16.26 & 11.33 & 17.73 \\
        TIN~\cite{interactiveness}               & 17.03 & 13.42 & 18.11 & 19.17 & 15.51 & 20.26 \\
        PMFNet~\cite{pmfnet}                     & 17.46 & 15.65 & 18.00 & 20.34 & 17.47 & 21.20 \\ 
        DJ-RN~\cite{djrn}                        & 21.34 & 18.53 & 22.18 & 23.69 & 20.64 & 24.60 \\ 
        \hline
        PPDM~\cite{ppdm}                         & 21.73 & 13.78 & 24.10 & 24.58 & 16.65 & 26.84 \\
        VCL~\cite{vcl}                           & 23.63 & 17.21 & 25.55 & 25.98 & 19.12 & 28.03 \\
        DRG~\cite{DRG}                           & 24.53 & 19.47 & 26.04 & 27.98 & 23.11 & 29.43 \\
        IDN~\cite{idn}                           & 26.29 & 22.61 & 27.39 & 28.24 & 24.47 & 29.37 \\
        Zou~\etal~\cite{zou2021_hoitrans}        & 26.61 & 19.15 & 28.84 & 29.13 & 20.98 & 31.57 \\
        ATL~\cite{hou2021atl}                    & 28.53 & 21.64 & 30.59 & 31.18 & 24.15 & 33.29 \\
        AS-Net~\cite{chen_2021_asnet}            & 28.87 & 24.25 & 30.25 & 31.74 & 27.07 & 33.14 \\
        QPIC~\cite{qpic}                         & 29.07 & 21.85 & 31.23 & 31.68 & 24.14 & 33.93 \\
        FCL~\cite{hou2021fcl}                    & 29.12 & 23.67 & 30.75 & 31.31 & 25.62 & 33.02 \\
        GGNet~\cite{zhong2021glance}             & 29.17 & 22.13 & 30.84 & 33.50 & 26.67 & 34.89 \\
        SCG~\cite{zhang:iccv2021}                & 31.33 & 24.72 & 33.31 & 34.37 & 27.18 & 36.52 \\
        CDN~\cite{cdn}                           & 31.78 & 27.55 & 33.05 & 34.53 & 29.73 & 35.96 \\
        \textbf{Ours}                            & \textbf{33.51} & \textbf{30.30} & \textbf{34.46} & \textbf{36.28} & \textbf{33.16} & \textbf{37.21} \\
        \hline
      \end{tabular}}}
      \vspace{-10px}
      \caption{Results on HICO-DET~\cite{hicodet}. The first part adopted COCO pre-trained detector. HICO-DET fine-tuned or one-stage detector is used in the second part. All the results are with \textbf{ResNet-50}.}
        \label{tab:hicodet} 
     \vspace{-10px}
\end{table}
\begin{table}[!t]
    \centering
    \resizebox{0.33\textwidth}{!}{\setlength{\tabcolsep}{0.6mm}{
      \begin{tabular}{l c c c}
         \hline
         Method         &$AP_{\text{role}}\text{(Scenario\ 1)}$  & $AP_{\text{role}}\text{(Scenario\ 2)}$ \\
         \hline
         \hline
         iCAN~\cite{gao2018ican}                   & 45.3 & 52.4 \\
         TIN~\cite{interactiveness}                & 47.8 & 54.2 \\
         VSGNet~\cite{vsgnet}                      & 51.8 & 57.0 \\
         IDN~\cite{idn}                            & 53.3 & 60.3 \\
         HOTR~\cite{hotr}                          & 55.2 & 64.4 \\
         QPIC~\cite{qpic}                          & 58.8 & 61.0 \\
         CDN~\cite{cdn}                            & 62.3 & 64.4 \\
         \hline
         \textbf{Ours}                           & \textbf{63.0} & \textbf{65.2} \\
         \hline
      \end{tabular}}}
      \vspace{-10px}
    \caption{Results with \textbf{ResNet-50} on V-COCO~\cite{vcoco}.}  
    \label{tab:vcoco}
    \vspace{-15px}
\end{table}
Table~\ref{tab:hicodet} compares our methods with previous state-of-the-art methods.
We outperform all of them with Default Full mAP of \textbf{33.51}.
Even compared with methods like ATL~\cite{hou2021atl} which adopted additional object attribute information, we achieve an impressive advantage of \textbf{4.98} mAP.
When comparing to other transformer-based methods such as HOTR~\cite{hotr}, \cite{zou2021_hoitrans}, AS-Net~\cite{chen_2021_asnet}, QPIC~\cite{qpic}, and CDN~\cite{cdn} our method manages to attain relative improvements of \textbf{30.2\%}, \textbf{16.1\%}, \textbf{15.3\%}, and \textbf{5.4\%}, respectively.
To fully verify the effectiveness of our method, we also adopt the very recent CDN~\cite{cdn} and outperform it significantly.
Note that even compared with CDN-L~\cite{cdn} (Default Full mAP 32.07) with more parameters, our model still maintains a significant advantage.

Table~\ref{tab:vcoco} compares our result on V-COCO~\cite{vcoco} with those of previous state-of-the-arts, which indicates that our method achieves impressive advantage over previous methods with \textbf{63.0} and \textbf{65.2} mAP under Scenario~1 and 2.

\noindent{\bf Results on Interactiveness Detection}
To better demonstrate our contribution to the H-O pair filtering, we evaluate our interactiveness detection~\cite{interactiveness} on HICO-DET~\cite{hicodet}.

First, following the interactiveness AP proposed in~\cite{interactiveness}, we evaluate our interactiveness detection, comparing with open-source state-of-the-arts~\cite{interactiveness,ppdm,qpic,cdn}.
In detail, we adopt $S_b$ as the interactiveness score for our model.
For TIN~\cite{interactiveness}, the inherent interactiveness score is adopted.
For PPDM~\cite{ppdm}, QPIC~\cite{qpic}, and CDN~\cite{cdn}, the mean of 520 HOI scores is used as an approximation.
Table~\ref{tab:bin_ap} tabulates the results, which shows the interactiveness AP of TIN is significantly lower, echoing our analysis that it suffers from the mass of exhaustively generated negative H-O pairs even with the non-interaction suppression~\cite{interactiveness}.
In terms of the one-stage PPDM~\cite{ppdm} directly detecting H-O pairs, the performances are better since the avoid of exhaustive pairing.
Surprisingly, the interactiveness performance gap between QPIC~\cite{qpic} and CDN~\cite{cdn} is negligible, while our method demonstrates to be considerably better than previous methods with interactiveness AP of \textbf{37.39}.

To verify that our method is superior on pair filtering, we select previous open-source state-of-the-arts and compare the Default Full mAP in a Top-k manner~\cite{ppdm} in Table~\ref{tab:topk}. That is, we only select the predictions with top-k confidence for each image. Even with only 5 predictions per image, the advantage is still impressive over other methods. 
\begin{table}[t]
\small
\centering
\resizebox{0.4\textwidth}{!}{
\begin{tabular}{c|@{\hspace{2mm}}c@{\hspace{2mm}}c@{\hspace{2mm}}c@{\hspace{2mm}}c|@{\hspace{2mm}}c@{\hspace{2mm}}}
        \hline
         & TIN~\cite{interactiveness} & PPDM~\cite{ppdm} &  QPIC~\cite{qpic} & CDN~\cite{cdn} & ours \\ \hline \hline
        AP & 14.35 & 27.34 & 32.96 & 33.55 & \textbf{37.39} \\
        \hline
\end{tabular}}
   \vspace{-10px}
    \caption{Interactiveness detection on HICO-DET~\cite{hicodet}.}
    \label{tab:bin_ap}
\vspace{1mm}
    \centering
    \resizebox{0.40\textwidth}{!}{
    \begin{tabular}{@{\hspace{6mm}}c@{\hspace{6mm}}|@{\hspace{6mm}}c@{\hspace{6mm}} c@{\hspace{6mm}} c@{\hspace{6mm}} c}
        \hline
        Methods & Top-5 & Top-10 & All \\
        \hline
        PPDM~\cite{ppdm} & 18.92 & 20.35 & 21.10 \\
        QPIC~\cite{qpic} & 29.07 & 29.29 & 29.07 \\
        CDN~\cite{cdn}   & 30.19 & 30.40 & 31.78 \\
        Ours             & \textbf{32.65} & \textbf{33.07} & \textbf{33.51} \\
        \hline
    \end{tabular}}
   \vspace{-10px}
    \caption{Top-K result on HICO-DET~\cite{hicodet}. ``All'' indicates Top-100 for PPDM~\cite{ppdm} and QPIC~\cite{qpic}, and Top-64 for CDN~\cite{cdn} and ours.}
    \label{tab:topk}
\vspace{1mm}
\centering
    \resizebox{0.4\textwidth}{!}{\setlength{\tabcolsep}{0.3mm}{
    \begin{tabular}{c|@{\hspace{7.5mm}}c@{\hspace{7.5mm}} c@{\hspace{7.5mm}} c@{\hspace{7.5mm}}}
        \hline
        Methods & Full & Rare & Non-Rare \\
        \hline
        \hline
        iCAN~\cite{gao2018ican}               & 14.16 & 12.26 & 14.73 \\
        iCAN~\cite{gao2018ican}$^{\rm{QPIC}}$ & 21.78 & 13.18 & 24.35 \\
        iCAN~\cite{gao2018ican}$^{\rm{CDN}}$  & 24.05 & 18.32 & 25.76 \\
        iCAN~\cite{gao2018ican}$^{\rm{Ours}}$ & \textbf{26.07} & \textbf{21.03} & \textbf{27.58} \\
        \hline
    \end{tabular}}}
    \vspace{-10px}
    \caption{Performance of iCAN~\cite{gao2018ican} on HICO-DET~\cite{hicodet} with different pair detection. Superscripts indicate the source of pair detection, where no superscript indicates the exhaustive pairing~\cite{gao2018ican}.}
    \label{tab:ican}
    \vspace{-15px}
\end{table}

Furthermore, we explore how our pair filtering can boost the performance of two-stage methods.
Following CDN~\cite{cdn}, we feed the representative two-stage method iCAN~\cite{gao2018ican} (using exhaustive pairing without pair filtering) with our detected pairs, and compare the result produced by feeding exhaustive pairs as input.
In addition, the results using CDN~\cite{cdn} and QPIC~\cite{qpic} pairs as input are also compared. Here, mAP under Default mode for the three sets (Full, Rare, Non-Rare) are reported.
Table~\ref{tab:ican} shows that the performance of iCAN is significantly boosted with the pairs of high-quality, especially of the ones from our method.

\subsection{Visualization}
\label{sec:vis}
\begin{figure}[!t]
    \centering
    \includegraphics[width=0.45\textwidth]{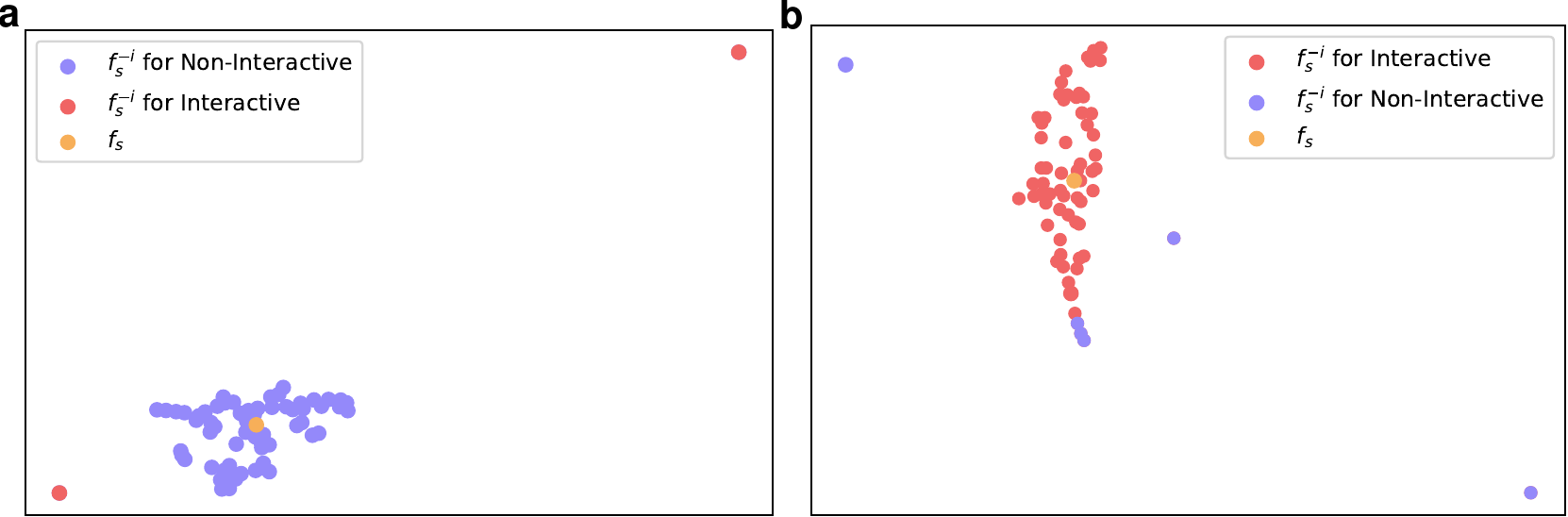}
    \vspace{-10px}    
    \caption{Field change visualization. $f_s$ (orange) is the field summary feature, while $f_s^{-i}$ of non-interactive pairs (purple) are in majority in the left; $f_s^{-i}$ of interactive pairs (red) are in majority in the right. As shown, $f_s^{-i}$ of minority pairs locates far from $f_s$.}
    \label{fig:vis}
    \vspace{-10px}
\end{figure}
Figure~\ref{fig:vis} visualizes the field change under the constraints (Section~\ref{sec:change2field}).
The field summary feature $f_s$ and the changed summary feature $f_s^{-i}$ of different pairs are visualized with t-SNE~\cite{tsne}, where $f_s^{-i}$ corresponding to minority pairs follow the constraints well, validating our design.
\subsection{Ablation Studies}
\label{sec:ablation}
We conduct ablation studies on HICO-DET~\cite{hicodet} under the Default mode, with the results in Table~\ref{tab:ablation}.
\begin{table}[!t]
    \centering
    \resizebox{0.4\textwidth}{!}{\setlength{\tabcolsep}{0.3mm}{  
    \begin{tabular}{@{\hspace{5mm}}c@{\hspace{5mm}}|@{\hspace{5mm}}c@{\hspace{5mm}} c@{\hspace{5mm}} c@{\hspace{5mm}}}
        \hline
           & Full & Rare & Non-Rare \\
        \hline
        \hline
        Ours                        & \textbf{33.51} & \textbf{30.30} & \textbf{34.46} \\    
        \hline
        w/o IFM                     & 30.54 & 26.04 & 31.88 \\
        w/o $S_b$                   & 33.30 & 29.76 & 34.35 \\
        \hline
        $g(\cdot)$ via FC           & 30.70 & 25.68 & 32.20 \\
        $g(\cdot)$ via clustering   & 30.97 & 26.86 & 32.20 \\
        cardinality only            & 32.38 & 27.99 & 33.69 \\
        field change only           & 32.76 & 28.82 & 33.94 \\
        \hline
        Unsup-IFM                   & 31.62 & 27.38 & 32.88 \\
        \hline
    \end{tabular}}}
    \vspace{-10px}
    \caption{Ablation studies on HICO-DET~\cite{hicodet}.}
    \label{tab:ablation}
    \vspace{-15px}
\end{table}

First, we show how the model is influenced if the interactiveness field module (IFM) is removed.
The considerable mAP drop of 2.97 validates the key role of IFM.
We then reveal the influence of interactive score $S_b$ on performance.
We find that removing $S_b$ only results in a minor drop.
This demonstrates that the IFM functions more than merely in results fusion: it also contributes to feature learning.

Second, different implementations of IFM are compared.
Replacing IFM with a fully-connected layer as done in TIN~\cite{interactiveness}, we obtain 30.70 mAP ($g(\cdot)$ via fully-connected in Table~\ref{tab:ablation}), which is slightly better than removing IFM while still insignificant.
By implementing $g(\cdot)$ via clustering as proposed in Section~\ref{sec:sys}, we achieve a marginal improvement compared to a model w/o IFM, far below the advanced version of $g(\cdot)$, showing the efficacy of our design.
This experiment on the other hand shows the importance of the \extrema prior even with a straightforward $g(\cdot)$ implementation.
Moreover, we evaluate the influence of different constraints.
With only cardinality constraint (Section~\ref{sec:rarity2field}), we suffer 1.13 mAP drop (cardinality only in Table~\ref{tab:ablation}).
While the mAP drop is 0.75 if only field change constraints (Section~\ref{sec:change2field}) are preserved (field change only in Table~\ref{tab:ablation}).

Third, we demonstrate the performance of IFM operating in the unsupervised mode, referred to as Unsup-IFM.
That is, we zero out the loss items proposed in Section~\ref{sec:field2bin}. 
Then, IFM is only restrained by the \extrema prior.
Even without supervision using interactiveness labels, we can achieve good improvement with only the \extrema prior.

Moreover, we validate IFM by the error between the number of predicted and GT interactive pairs per image of different implementation of $g(\cdot)$. The predicted interactive pair number is calculated by summing the predicted interactive probability of each pair.
The results in Table~\ref{tab:error} show the advanced implementation does exploit the prior. 
The impressive gap with and w/o IFM proves that the raw \textit{data-driven} methods fail to model the bimodal distribution well.
\begin{table}[!t]
    \centering
    \vspace{-10px}
    \resizebox{0.8\linewidth}{!}{\setlength{\tabcolsep}{0.8mm}{\begin{tabular}{c|c c c}
        \hline
        Dataset       & $\frac{\#inter}{\#no-inter} \ll 1$ & $\frac{\#inter}{\#no-inter} \approx 1$ & $\frac{\#inter}{\#no-inter} \gg 1$ \\
        \hline
        w/o IFM                   & 0.38 & 0.55 & 2.34 \\
        $g(\cdot)$ via FC         & 0.32 & 0.57 & 2.12 \\
        $g(\cdot)$ via clustering & 0.28 & 0.51 & 2.09 \\
        Ours                      & \textbf{0.19} & \textbf{0.42} & \textbf{1.88} \\
        \hline
    \end{tabular}}}
    \vspace{-10px}
    \caption{Error of \#interactive pairs between prediction and GT.}
    \label{tab:error}
    \vspace{-20px}
\end{table}

\vspace{-10px}
Finally, we demonstrate the performance under different interactive ratios. The IFM brings relative improvement as 9.23\% (30.68 to 33.52), 0.11\% (52.98 to 53.04), 3.04\% (51.42 to 52.98), respectively with interactive ratio $\ll 1, \approx 1, \gg 1$. These show our impressive improvement upon valid cases and ignorable harm on invalid cases.
For more limitation and social impact discussion, please refer to the appendices.

\section{Conclusion}
This paper focuses on previously overlooked interactiveness \extrema prior in HOI learning.
To utilize this prior, the interactiveness field is proposed and modeled.
Multiple properties of the proposed field are explored to match the learned field and realistic HOI scenes.
Our method effectively discriminates interactive human-object pairs and achieves significant improvements, validated on widely-used benchmarks. 
Though interactiveness field prompts H-O pairing and boosts HOI detection, we believe the room for H-O pairing is still large and needs more explorations.

{\small
\bibliographystyle{ieee_fullname}
\bibliography{egbib}
}

\clearpage
\appendix
\setcounter{figure}{0}
\setcounter{table}{0}
\appendixpage
\begin{strip}
\centering\includegraphics[width=0.95\textwidth]{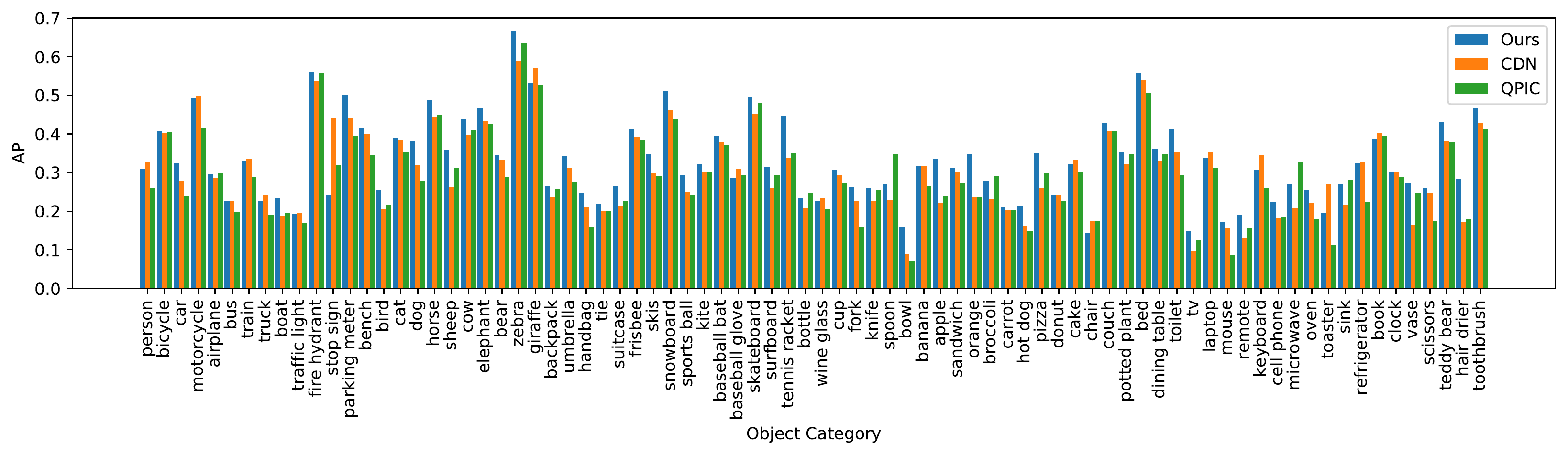}
\captionof{figure}{Interactiveness AP for different object categories of our model, QPIC~\cite{qpic}, and CDN~\cite{cdn}.}
\label{fig:inter_ap}
\vspace{-15px}
\end{strip}

\begin{strip}
\centering\includegraphics[width=0.95\textwidth]{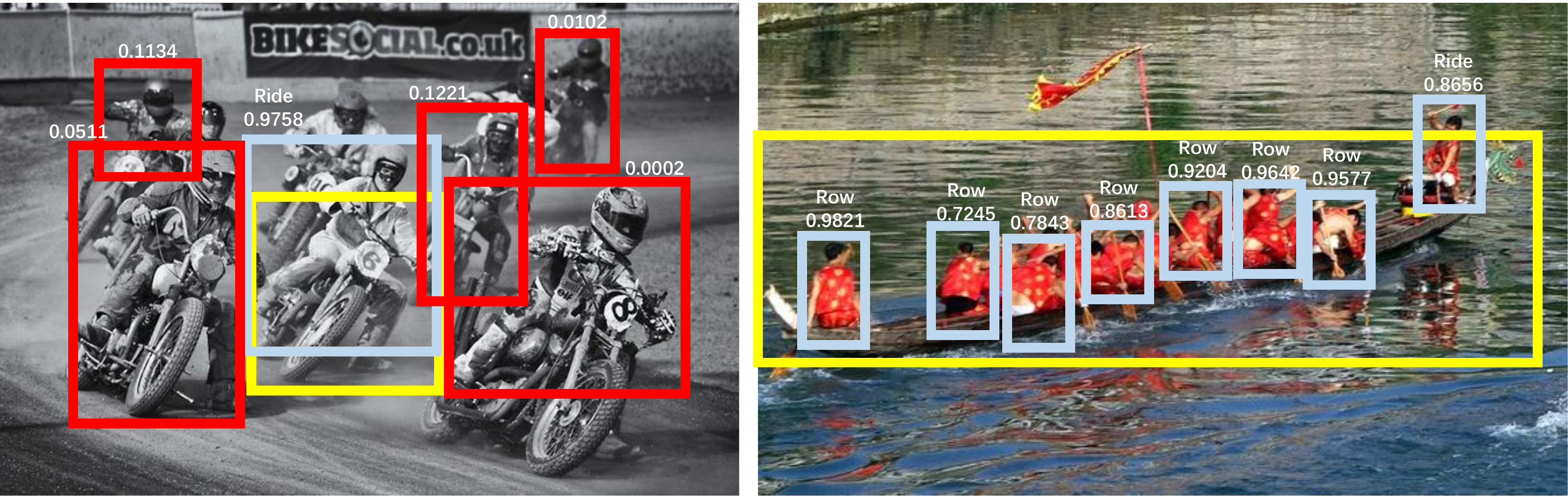}
\captionof{figure}{Some typical detection results on HICO-DET~\cite{hicodet}. Given an object (yellow), our model efficiently pairs the object with the interactive human (cerulean) and filters out the non-interactive human instances (red). Under both circumstances (non-interactive pairs in majority on the left, and interactive pairs in majority on the right), we achieve satisfactory results.}
\label{fig:hardcase}
\end{strip}

\section{Generalization of Interactiveness Bi-modal Prior}

A potential issue is the generalization of our proposed bi-modal prior. 
We reemphasize that the bimodal prior is universal with good generalization in the context of HOI in two aspects.
\textbf{First}, 
the bimodal prior is fundamentally rooted in HOI. 
The very compositional nature of human and object in HOI makes it susceptible to a severely imbalanced distribution as revealed by \textit{Zipf's Law}.
As shown in Table~\ref{tab:ratio}, widely used natural image HOI datasets~\cite{hicodet,vcoco,djrn,AVA} all hold the prior
\begin{table}[!ht]
    \centering
    \vspace{-10px}
    \resizebox{0.8\linewidth}{!}{\setlength{\tabcolsep}{0.8mm}{\begin{tabular}{c|c c c}
        \hline
        Dataset       & $\frac{\#inter}{\#no-inter} \ll 1$ & $\frac{\#inter}{\#no-inter} \approx 1$ & $\frac{\#inter}{\#no-inter} \gg 1$ \\
        \hline
        HICO-DET~\cite{hicodet}      & 79.1\% & 7.3\% & 13.6\% \\
        V-COCO~\cite{vcoco}        & 76.0\% & 7.9\% & 16.1\% \\
        Ambiguous-HOI~\cite{djrn} & 80.1\% & 6.2\% & 13.7\% \\
        AVA~\cite{AVA}           & 73.2\% & 8.4\% & 18.4\% \\
        \hline
    \end{tabular}}}
    \vspace{-10px}
    \caption{Interactive ratio of different datasets.}
    \label{tab:ratio}
    \vspace{-18px}
\end{table}
\textbf{Second}, we claim that the \textbf{object-centric} bimodal prior exploited in our paper is one subclass of the prior, since the widely-used benchmarks HICO-DET and V-COCO both have this property.
Besides the object-centric prior that is more suitable in \textit{multi-person} scene, a similar prior exists in a \textbf{human-centric} view for images with \textit{few people}.
Even for really sparse scenes containing one person and one object, in a \textbf{body-part view} inspired by \citeappend{partstate}, the interactive body parts are statistically rare.
For such sparse scenes, 
statistics show in images with only \textit{one person} and \textit{one object} from HAKE~\citeappend{hake_pami,li2019hake} that only \textbf{9.8}\% of the existing parts are interactive with objects. 
That said, the prior still holds as a \textit{learning paradigm}.
We believe the object-centric prior is a first step towards deeper exploration on such useful prior.

\section{Detailed Analysis on Interactiveness \\Detection}
As stated in the main paper in Section 4.3, we evaluate our model using the interactive AP metric proposed by TIN~\cite{interactiveness}.
In this section, we include more details for interactiveness detection.
Figure~\ref{fig:inter_ap} shows the interactiveness AP for different object categories of our model and previous state-of-the-art QPIC~\cite{qpic} and CDN~\cite{cdn}. 
Our model achieves superior performance on most of the object categories.
In detail, our method takes the lead in \textbf{56} of the 80 object categories, while falling behind on only 4 categories.
Furthermore, on over 20 object categories, our advantage is more than 5 mAP, indicating the efficacy of our interactiveness detection for various objects.

\section{Prediction Visualization}
To vividly show the effectiveness of our method, we give some typical results on HICO-DET~\cite{hicodet} in Figure~\ref{fig:hardcase}. Our method can precisely filter out the non-interactive pairs while detecting interactive pairs in complex scenes.

\section{Discussion on Limitations}

Though the interactiveness field has greatly enhanced H-O pairing and boosted the HOI detection, the room for H-O pairing is still large needing more exploration.

While the proposed bimodal prior is of great efficacy in interactiveness modeling, it is still an issue to precisely discern the correspondence between rare/non-rare pairs and interactive/non-interactive pairs.
Since even with a compromised strategy that treats rare pairs as interactive, the performance improvement is considerable, we believe effective inference on the correspondence may lead to very promising enhancement.

The proposed interactiveness field is investigated generally based on the bimodal prior only, while we believe the more fine-grained study is worthwhile, e.g., the interactiveness field for different object categories inspired by Liu~\etal~\citeappend{liu2022highlighting}, the interactiveness field for different verb categories, the field under different background contexts, the interplay of the interactiveness fields of different objects, and so on.

\section{Societal Impact}
As all the data used here come from public dataset thus there is no privacy issue. Our work aims at prompting the HOI understanding, thus may be helpful to the development of health-care robot, etc. However, there could be potentially negative societal implications, such as its potential use in surveillance, military purposes which requires serious moral consideration. We encourage well-intended application of our method.

\section{Licenses of Adopted Datasets}
V-COCO~\cite{vcoco} is released under the MIT License.
Our code is mostly derived from DETR~\cite{detr}, QPIC~\cite{qpic} and TIN~\cite{interactiveness}.
DETR~\cite{detr} and QPIC~\cite{qpic} are released under the Apache License 2.0.
While TIN~\cite{interactiveness} is released under the MIT License.

{\small
\bibliographystyleappend{ieee_fullname}
\bibliographyappend{egbib}
}

\end{document}